# Faster than real-time detection of shot boundaries, sampling structure and dynamic keyframes in video

Hannes Fassold
*DIGITAL*
*JOANNEUM RESEARCH*
Graz, Austria
hannes.fassold@joanneum.at

*Abstract*— The detection of shot boundaries (hardcuts and short dissolves), sampling structure (progressive / interlaced / pulldown) and dynamic keyframes in a video are fundamental video analysis tasks which have to be done before any further high-level analysis tasks. We present a novel algorithm which does all these analysis tasks in an unified way, by utilizing a combination of inter-frame and intra-frame measures derived from the motion field and normalized cross correlation. The algorithm runs four times faster than real-time due to sparse and selective calculation of these measures. An initial evaluation furthermore shows that the proposed algorithm is extremely robust even for challenging content showing large camera or object motion, flashlights, flicker or low contrast / noise.

*Keywords*— video analysis, shot detection, sampling structure detection, keyframe detection, real-time analysis

## I. INTRODUCTION

In order to perform high-level computer vision tasks (like object detection and tracking, restoration and enhancement etc.) on video content, first some fundamental preprocessing tasks have to be performed in advance. Specifically, the content has to be split into individual shots (shot boundary detection), which are usually separated by hardcuts or short dissolves.

It is also crucial to detect the sampling structure of the video. The sampling structure can be progressive, interlaced (each frame contains two fields of half width which are from consecutive timepoints) or 3:2 pulldown (the standard method for converting progressive film content with 24 frames per second to interlaced video content with 60 fields per second). For example, for an interlaced video it is not advisable to use the whole frame, as it will exhibit combing artifacts if motion is present in the scene (as the two fields are from different timepoints). For interlaced video, the information about the field order (either upper field first or lower field first) is also desired.

Finally, for extracting an image dataset for training neural networks from an video an algorithm for the extraction of dynamic keyframes is needed. Dynamic keyframes are non-uniformly spaced frames which adapt to the variation in the video. In video segments with high variation (e.g. fast motion scene) more keyframes will be extracted, whereas in a static video segment the spacing between the keyframe will be much larger. In order to be able to process large video collections (e.g. broadcaster archives can have video collections comprising several hundreds of thousands of hours) it is crucial that the preprocessing is done as fast as possible, at least multiple times faster than real-time is desired usually.

Addressing this, we propose a novel method which does shot detection, sampling structure detection and dynamic keyframe extraction in an unified way. Due to the unified approach and sparse and selective calculation of the content-based measures, it is able to run four times faster than real-time for videos in 2K resolution.

The remainder of the work is organized as follows. In section 2 we revise related work. Section 3 presents the proposed algorithm. Section 4 describes our initial evaluation of the algorithm with respect to quality and runtime. Section 5 gives some information about the demo application we developed for showcasing the algorithm. Finally, section 6 concludes the paper.

## II. RELATED WORK

Up to our knowledge, there is no method currently in the literature which combines all three tasks (shot boundary detection, sampling structure detection, dynamic keyframe extraction) in an unified way in *one* algorithm, as we do. In the following, we therefore present methods which do only a single task. In order to match the functionality of our algorithm, these methods would have to be run sequentially for a video without re-using intermediate features, which comes of course with a disadvantage in runtime.

A variety of algorithms have been proposed in the literature for shot boundary detection, either with classical approaches using hand-crafted features or with neural networks..The authors of [1] proposed a wavelet based feature vector that measures four main quantities: color, edge, texture, and motion strength. Features are calculated for each frame and shot boundaries are deducted from the temporal evolution of these features. The *DeepSBD* algorithm [2] pioneered the use of a neural networks for shot boundary detection. It employs a network which consists of five 3D convolutional layers which is trained on a large dataset created by the authors of this work. They utilize segments of 16 frames with 8 frames overlap and therefore do *not* provide frame-level accuracy, in contrast to our proposed algorithm which is able to provide accurate shot boundaries. The *TransNetV2* algorithm from [4] integrates various techniques such as convolution kernel factorization, batch normalization and skip connections, resulting in improved performance. The recent work of [5] proposes the *AutoShot* algorithm. They derive the architecture of the employed neural network by conducting neural architecture search in a search space encapsulating various advanced 3D ConvNets and Transformer.

For sampling structure detection only a few works have been proposed in the literature, although there are also some patents which have been filed. The authors of [6] argument that due to interlacing artifacts in the full frame in the presence of motion, a difference in isophote curvature can be measured and a threshold for effective classification can be set. They utilize two different measures in their algorithm, namely Kullback-Leibler divergence and Canny edge detection. The algorithm from [7] counts the number of zipper artifacts of a certain minimum length in the top field and the bottom field between two frames and bases the decision on this measure. Note that both algorithms [6] and [7] are only able to detect progressive or interlaced, whereas our algorithm can detect also 3:2 pulldown and the field order of interlaced content.

For dynamic keyframe extraction, we refer to the survey given in [3] which provides a good overview about state of the art methods for keyframe extraction.

### III. ALGORITHM

In contrast to many detection algorithms from the literature which use a large number of diverse image features and a classifier (like a neural network or support vector machine) which is trained on some ground truth, we employ only a few features and use a hand-crafted approach. This does not have to be a disadvantage, as the features we are using are highly discriminative. And by not using a trained classifier, we are better able to address potential algorithmic issues on difficult content during research and development of the algorithm.

#### A. Key features

The choice of the employed features is motivated by the following observations:

- The image content in temporally neighbouring frames in the same shot is usually very similar, and should be identical when camera and object motion is compensated properly. The calculated motion field (with a robust optical flow algorithm) between two neighbouring frames should express the camera and object motion and therefore should be regular and not have too high magnitude (except for high motion scenes).

- The image content in two frames from different shots is very different. Furthermore, when calculating the motion field between these two frames, the optical flow algorithm will get in trouble as it is not able to find proper pixel matches. Usually, this means that the calculated motion field will be very irregular, and the magnitude of the motion vectors is very high. This behaviour can be easily seen in the example given in te left side of Fig. 1.

Based on these observations, we are employing the following features:

- $H(t)$ is the intensity histogram of the image at timepoint $t$. Various measures (mean, standard deviation, mean absolute deviation) are calculated from the histogram.

- $AMM(I, J)$ is the average magnitude of the motion vectors of the motion field calculated between the images $I$ and $J$. For calculating the motion field, we employ the Dense Inverse Search optical flow algorithm from [8]. It runs very fast employing only the CPU and is quite robust against brightness variations. It works well also for large motion in the scene (e.g. sports videos), which is especially important for the shot detection component.

- $SWR(I, J)$ is the image dissimilarity between the reference image $I$ and the warped (motion-compensated) $J$. The warped image is generated by calculating the motion field between both images and warping the image J with the motion field. When the motion compensation works properly, then the warped image should be identical to the reference image. We employ the normalised cross correlation (*NCC*) similarity measure in order to be invariant against brightness variations due to flicker or camera flashlights.

- $ACT(I, J)$ is the geometric average of $AMM(I, J)$ and $SWR(I, J)$ and measures the activity between the images $I$ and $J$. If the images are very similar and there is not a lot a motion between them, $ACT(I, J)$ will be nearly zero, whereas in the opposite case its value will be high.

The activity between consecutive video frames $ACT(I_t, I_{t+1})$ is calculated always. All other measures $ACT(I_t, I_s)$ are calculated sparsely and selectively, only if they are beneficial to verify a certain hypothesis (e.g. the hypothesis that the current shot is interlaced).

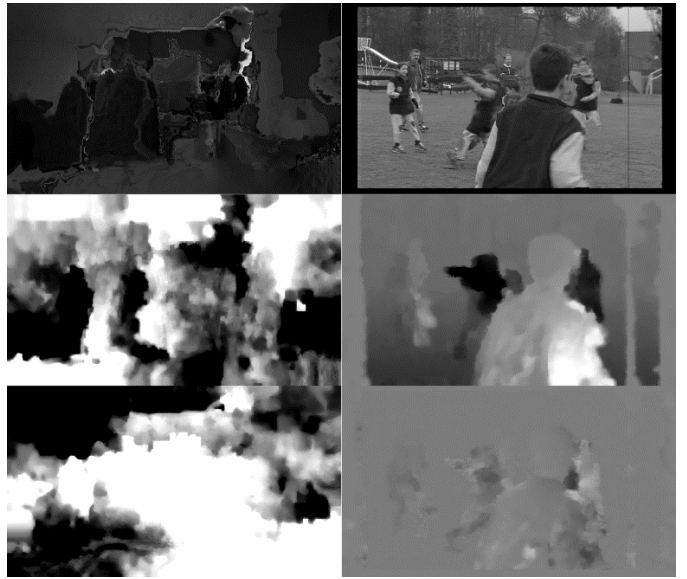

Fig. 1. Left column: warped image and motion field for two frames from *different* shots. Right column: warped image and motion field for two frames for the *same* shot. For the motion field, the x-disparity part is visualized in the second row and the y-disparty part in the last row. Disparity is visualized in grayscale (zero value = gray, positive value = white, negative value = black). One can see that the calculated motion field between frames of *different* shots has large disparity values and is very irregular, and the warped image is very ragged and does not look like normal video content.

For this purpose we implemented a dedicated software framework where measures are calculated *on-demand and cached*, in order to ensure that two algorithm components (e.g. shot detector and dynamic keyframes detector) do not calculate the same measure twice. It works for unary measures (with only one frame as input – e.g. histogram) as well as binary measures (having two frames as input – e.g. optical flow) and global measures.

*B. Shot boundary detection*

The shot detector comprises two phases. The first phase (fast check) does for each frame a check whether it is possible that at this frame a hardcut or short dissolve (consisting of up to 4 frames) occurs. As the first phase is done for every frame, it must be very fast. The second phase (deep check) is only invoked if the first phase decides that at this frame it is possible that a short dissolve occurs. It tests for each $K \in 1...4$ whether the hypothesis of a *K*-frame dissolve at this frame is valid or not (note that $K = 1$ denotes a hardcut). The second phase is not invoked very often and therefore can be computationally much more expensive without impacting the overall runtime negatively.

A hypothesis for a $K-$ frame dissolve is verified if the activity $ACT(I_t, I_{t-j})$ between the last frame $I_t$ before the dissolve and its predecessors $I_{t-j}$ is significantly smaller than the activity $ACT(I_t, I_{t+K})$ between the last frame before the dissolve and the first frame after it. This makes sense, as the activity $ACT(I_t, I_{t+K})$ will be high if the frames $I_t$ and $I_{t+K}$ are from different shots.

*C. Sampling structure detection*

The sampling structure detector utilizes a combination of inter-frame and intra-frame activity measures. Each frame It is split into its upper field $I_{t,u}$ and lower field $I_{t,l}$, then we calculate the three basic measures $v0 = ACT(I_{t,u}, I_{t,l})$, $v1 = ACT(I_{t,u}, I_{t+1,l})$ and $v2 = ACT(I_{t,l}, I_{t+1,u})$. Each sampling structure type has now a very characteristic pattern in the relation of the measures $v0$, $v1$ and $v2$.

Progressive content is characterized by a near-zero value of $v0$, whereas the values $v1$ and $v2$ are non-zero and approximately equal. Interlaced content is characterized by nonzero values of $v0$, $v1$ and $v2$. Furthermore, the values $v1$ and $v2$ are not approximately equal for interlaced content, because one corresponds to fields which are significantly further apart in time. When interlaced content has been detected, additionally the field order (upper field first vs. lower field first) is also determined. We employ the ratio $\beta = \frac{v1}{v2}$ for this purpose, whose value will be smaller than 1 for field order 'lower field first' and bigger than 1 for field order 'upper field first'. For 3 : 2 pulldown, the pattern is significantly more complex, as it depends also on the position of the frame within a 'pulldown unit' consisting of 5 frames.

By analyzing several frames of the shot statistically, we can set up a hypothesis now whether its sampling structure is progressive, interlaced or 3 : 2 pulldown. Usually, it is enough to analyze at most 15 to 20 frames of the shot. We skip static frames with no apparent inter-frame motion between current and next frame in the analysis, as no information can be gathered from those.

*D. Dynamic keyframe extraction*

The principle of the dynamic keyframe detector is straightforward. Within a shot, we are accumulating the activity values $ACT(I_t, I_{t+q})$ between consecutive frames. If the accumulated sum is higher than a certain threshold, then we trigger a keyframe for the current frame and set the accumulated sum back to zero.

## IV. EVALUATION

An initial evaluation has been done of the algorithm with respect to quality (detection capability, robustness, false positives) and runtime. Regarding runtime, the detector is able to process 2K (2048 x 1536) content roughly four times faster than real-time (~ 11 milliseconds per frame). For 4K video content, the algorithm is roughly three times faster than real-time (~ 14 milliseconds per frame). The detector implementation uses multiple CPU threads (4 CPU threads), but it does not employ GPU acceleration currently.

Regarding quality, the evaluation shows that the developed shot boundary detector algorithm is extremely robust even for challenging content with arge camera or object motion, flashlights, flicker, low contrast and the like. A major reason for the robustness of the algorithm is likely the usage of motion compensation backed by a high-quality optical flow algorithm and of a brighness-invariant similarity measure (normalized cross correlation). Of course, in extreme cases of motion or very heavy flicker even our highly robust employed measures might have issues, but our method is explicitly designed to be much more robust to motion and brightness variations than most algorithms in the literature.

In Fig. 2, some examples for correctly detected short dissolves and hard cuts in challenging content are given.

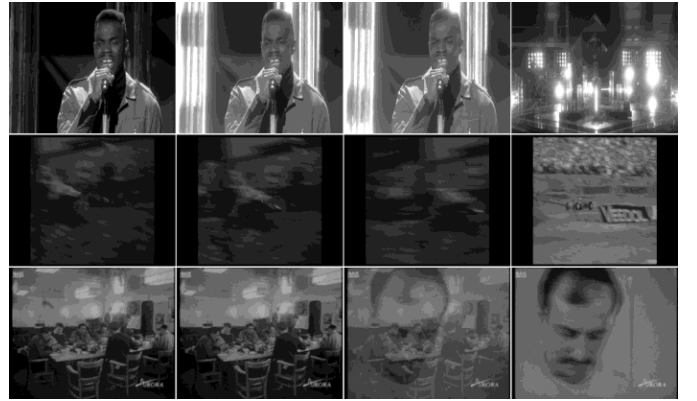

Fig. 2. Some examples for successfully detected short dissolves and hard cuts in challenging content (flicker, fast motion, motion blur …). The first three columns show the last three frames of the current shot, and the last column shows the first frame of the next shot.

A qualitative evaluation of the sampling structure detector on diverse progressive, interlaced and pulldown content shows that the algorithm is able to detect reliably the sampling structure as well as the field order (for interlaced content). Due to the usage of the same robust features like the shot detector, it is also very robust against fast camera or object motion, brightness variations, noise, low contrast and the like. It is able to detect the correct sampling structure also for video content for which it is difficult to discern whether the content is progressive or interlaced due to low amount of motion present in the scene. One example for this type of content are videos from weather panorama cameras, which usually have only minimal horizontal camera panning and often are also of low contrast due to cloudy weather or fog.

Finally, a qualitative evaluation of the dynamic keyframe detector shows that it adapts very well to the dynamic present in the video. So for video content where this is a high amount of motion present (like sports videos), it extracts keyframes in shorter intervals, whereas for content with low motion it extracts the keyframes in larger intervals. Typically, a keyframe is extracted every 8 – 30 frames.

## V. DEMO APPLICATION

We have developed also a demo application for the demonstration of the capabilities of the algorithm. The demo application is a mixed C++ / Python application, which runs on a Intel PC. The core algorithm has been implemented in C++ and runs solely on the CPU. In order to make the algorithm available in Python, a Python wrapper has been developed for it with the *pybind11* package.

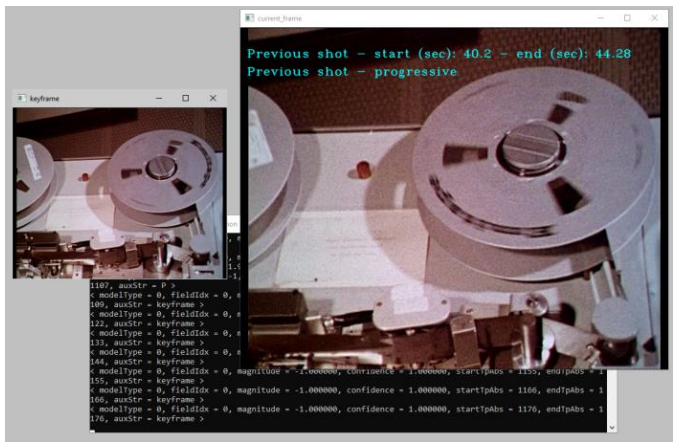

Fig. 3. Screenshot of the demo application.

The demo application (see Fig. 3) opens a video, runs the algorithm on it and shows the detection results. The dynamic keyframes are shown in a separate window as soon as they are detected. The result of the shot and sampling structure detector (progressive / interlaced / pulldown) for the previous shot is overlaid as text onto the window showing the input video. Naturally, the shot boundary can be only detected for the previous shot, as for the current shot it is still to be determined.

The algorithm runs multiple times faster than real-time, but the playback in the demo application has been slowed down so that the video is shown in normal playing speed.

A download link to the demo video is given in the Google Cloud at https://drive.google.com/file/d/17eFQeMtCusaQZjEb9wf3JX0qyAtrI8Rs/view?usp=sharing .

## VI. CONCLUSION AND FUTURE WORK

Identifying shot boundaries, analyzing the sampling structure and detecting dynamic keyframes are essential preliminary steps in video analysis. We have developed a novel algorithm that integrates these tasks into a single unified framework. The algorithm leverages a combination of inter-frame and intra-frame metrics, derived from motion field analysis and normalized cross correlation, to achieve accurate results.

Notably, our approach is computationally efficient, running four times faster than real-time, thanks to its sparse and selective calculation of these metrics. Initial testing has demonstrated a high robustness of the algorithm even for challenging video content featuring rapid camera or object movement, flashing lights, flicker, or low contrast and noise.

In the future, we will do a quantitative evaluation of the algorithm on popular evaluation datasets for shot boundary detection, sampling type detection and dynamic keyframe extraction.

ACKNOWLEDGMENT

This work was supported by European Union´s Horizon 2020 research and innovation programme under grant number 951911 - AI4Media.

REFERENCES

[1] L. Priya and D. S., "Walsh hadamard transform kernel-based feature vector for shot boundary detection," IEEE Transactions on Image Processing (TIP), vol. 23, no. 12, pp. 5187–5197, 2014.

[2] Ahmed Hassanien, Mohamed Elgharib, Ahmed Selim, Sung-Ho Bae, Mohamed Hefeeda, and Wojciech Matusik, "Large-scale, fast and accurate shot boundary detection through spatio-temporal convolutional neural networks". arXiv preprint arXiv:1705.03281, unpublished, 2017.

[3] Sujatha C. and Mudenagudi U., "A Study on Keyframe Extraction Methods for Video Summary", ICCICS, 2011.

[4] Tomas Soucek and Jakub Loko, "Transnet v2: An effective deep network architecture for fast shot transition detection", arXiv preprint arXiv:2008.04838, unpublished, 2020.

[5] Wentao Zhu, Yufang Huang and Xiufeng Xie, "AutoShot: A Short Video Dataset and State-of-the-Art Shot Boundary Detection", CVPR, 2023.

[6] Sune Hogild Keller, Kim Steenstrup Pedersen, and Francois Lauze, "Detecting interlaced or progressive source of video," in MMSP, 2005.

[7] Todd R. Goodall and Alan C. Boviki, "Detecting Source Video Artifacts with Supervised Sparse Filters", Picture Coding Symposium, 2018.a

[8] Till Kroeger and Radu Timofte, "Fast optical flow using dense inverse search," in ECCV 2016, 2016.